\documentclass{article}
\usepackage{spconf,amssymb,amsmath,epsfig}
\usepackage{microtype,url}
\usepackage{graphicx}
\usepackage{multirow}
\usepackage{booktabs}
\usepackage{amssymb,amsmath,graphicx}
\usepackage{caption}
\usepackage{subcaption}
\usepackage[export]{adjustbox}
\usepackage{soul}
\usepackage{moredefs} 
\usepackage{xcolor}
\usepackage{tipa}
\definecolor{darkspringgreen}{rgb}{0.09, 0.45, 0.27}

\defcommand{\vec}[1]{\mathbf{#1}} 

\def\vx{\mathbf{x}}

\def\vy{\mathbf{y}}

\def\ve{\mathbf{e}}

\def\eos{\texttt{</s>} }

\title{A Streaming On-Device End-to-End Model Surpassing Server-Side Conventional Model Quality and Latency}
\name{Tara N. Sainath\textsuperscript{*}, Yanzhang He\sthanks{Equal contribution}, Bo Li, Arun Narayanan, Ruoming Pang, Antoine Bruguier, Shuo-yiin Chang,}
\nameplus{Wei Li, Raziel Alvarez, Zhifeng Chen, Chung-Cheng Chiu, David Garcia, Alex Gruenstein, Ke Hu, Minho Jin,}
\nameplusplus{Anjuli Kannan, Qiao Liang, Ian McGraw, Cal Peyser, Rohit Prabhavalkar, Golan Pundak, David Rybach,}
\nameplusplusplus{Yuan Shangguan, Yash Sheth, Trevor Strohman, Mirk\'{o} Visontai, Yonghui Wu, Yu Zhang, Ding Zhao}
\address{Google, LLC, USA \\
\fontsize{9}{9}\selectfont\ttfamily\upshape
\{tsainath, yanzhanghe\}@google.com}

\begin{document}
\maketitle
\ninept
\begin{abstract}
Thus far, end-to-end (E2E) models have not been shown to outperform state-of-the-art
conventional models with respect to both quality, i.e., word error rate (WER), and latency,
i.e., the time the hypothesis is finalized after the user stops speaking. In this paper,
we develop a first-pass Recurrent Neural Network Transducer (RNN-T) model and
a second-pass Listen, Attend, Spell (LAS) rescorer that surpasses a conventional
model in both quality and latency. On the quality side, we incorporate a large
number of utterances across varied domains \cite{Arun19} to increase acoustic diversity
and the vocabulary seen by the model. We also train with accented English speech to
make the model more robust to different pronunciations. In addition, given the
increased amount of training data, we explore a varied learning rate schedule.
On the latency front, we explore using the end-of-sentence decision emitted by
the RNN-T model to close the microphone, and also introduce various optimizations to
improve the speed of LAS rescoring. Overall, we find that RNN-T+LAS offers a
better WER and latency tradeoff compared to a conventional model. For example,
for the same latency, RNN-T+LAS obtains a 8\% relative improvement in WER, while
being more than 400-times smaller in model size.
\end{abstract}

\section{Introduction \label{sec:introduction}}

End-to-end (E2E) models~\cite{Ryan19,CC18,Graves12, GravesMohamedHinton13,RaoSakPrabhavalkar17,Chan15, KimHoriWatanabe17,ChiuRaffel17} have gained large popularity in the automatic speech recognition (ASR) community over the last few years. These models replace components of a conventional ASR system, namely an acoustic (AM), pronunciation (PM) and language models (LM), with a single neural network. These models are a fraction of the size of a conventional ASR system, making them attractive for on-device ASR applications.
Specifically, on-device means that instead of streaming audio from the device to the server, recognizing text on the server, and then streaming results back to the device, recognition is performed entirely on the device. This has important implications for reliability, privacy and latency.

Running an ASR model on-device presents numerous additional user interaction constraints. First, we require that recognition results be streaming; the recognized words should appear on the screen as they are spoken. Second, the delay between when a user stops speaking and the hypothesis is finalized, which we refer to as latency, must be low. RNN-T models, which meet these on-device constrains, have been shown to be competitive in terms of quality in recent studies \cite{Ryan19,Arun19}. But under low-latency constrains, they lag behind a conventional server-side streaming ASR system~\cite{Ryan19}. At the other end of the spectrum, non-streaming models, such as LAS, have been shown to outperform a conventional ASR system~\cite{CC18}. However, LAS models are not streaming as they must attend to the entire audio segment. Recently, a 2-pass RNN-T+LAS model was proposed in \cite{SainathPang19}, where LAS rescores hypotheses from RNN-T. This model was shown to abide by user interaction constraints, and offer comparable performance to a conventional model.

In this paper, we extend on the work from \cite{SainathPang19} in several directions, to develop an on-device E2E model that surpasses a conventional model \cite{Golan16} in both WER and latency. First, on the quality-front, we train our model on multi-domain audio-text utterance pairs, utilizing sources from different domains including search traffic, telephony data and YouTube data \cite{Arun19}. This not only increases acoustic diversity, but also increases the vocabulary seen by the E2E model, as it is trained solely on audio-text pairs which is a small fraction compared to the text-only LM data used by a conventional model. Because the transcription and audio characteristics vary between domains, we also explore adding the domain-id as an input to the model. We find that by training with multi-domain data and feeding in a domain-id, we are able to improve upon a model trained on voice search data only. Second, also on the quality-front, we address improving robustness to different pronunciations. Conventional models handle this by using a lexicon that can have multiple pronunciations for a word. Since our E2E models directly predict word-pieces \cite{Schuster2012}, we address this by including accented English data from different locales \cite{li2018multi}. Third, given the increased audio-text pairs used in training, we explore using a constant learning rate rather than gradually decaying the learning rate over time, thereby giving even weight to the training examples as training progresses.

We also explore various ideas to improve latency of our model. We define \emph{endpointer (EP) latency} as the amount of time it takes for the microphone to close after a user stops speaking. To make a fair comparison, this metric excludes network latency and computation time when comparing the on-device and server endpointer latencies. Typically, an external voice activity detector (VAD) is used to make microphone-closing decisions. For conventional ASR systems, an end-of-query (EOQ) endpointer \cite{Matt17,chang2017endpoint,chang19unified} is often used for improved EP latency. Recently, integrating the EOQ endpointer into the E2E model by predicting the end-of-query symbol, \eos, to aid in closing the microphone was shown to improve latency \cite{Shuoyiin19}. We build on this work here, introducing a penalty in RNN-T training for emitting \eos too early or too late. Second, we improve the \emph{computation latency} of the 2nd-pass rescoring model. Specifically, we reduce the 2nd-pass run time of LAS by batching inference over multiple arcs of a rescoring lattice, and also offloading part of the computation to the first pass. LAS rescoring also obtains better tradeoff between WER and EP latency due to the improved recognition quality.


\section{Model Architecture}
\label{sec:model}

\begin{figure}[h!]
  \centering
  \includegraphics[scale=0.4]{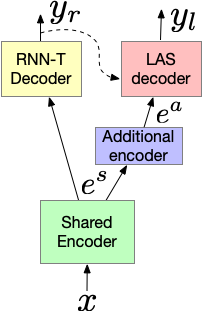}
  \caption{{Two-Pass Architecture}}
   \label{fig:2pass_architecture}
   \vspace{-0.2in}
\end{figure}

The proposed 2-pass E2E architecture \cite{SainathPang19} is shown in Figure \ref{fig:2pass_architecture}. Let us denote input acoustic frames as  $\vx=(\vx_1 \ldots \vx_T)$, where $\vx_t \in \mathbb{R}^d$ are stacked log-mel filterbank energies ($d=512$) and $T$ the number of frames in $\vx$. In the 1st-pass, each acoustic frame $\vx_t$ is passed through a shared encoder, consisting of a multi-layer LSTM, to get output $\ve^s_t$, which is then passed to an RNN-T decoder~\footnote{RNN-T decoder consists of a prediction network and a joint network.} that predicts $\vy_r=\{y_1, \ldots y_T\}$, the output sequence, in a streaming fashion. Here $\vy_r$ is a sequence of word-piece tokens \cite{Schuster97}. In the 2nd-pass, the full output of the shared encoder, $\ve^s=(\ve^s_1 \ldots \ve^s_T)$, is passed to a small additional encoder to generate $\ve^a=(\ve^a_1 \ldots \ve^a_T)$, which is then passed to an LAS decoder. We add the additional encoder since it is found to be useful to adapt the encoder output to be more suitable for LAS.
During training, the LAS decoder computes output $\vy_l$ according to $\ve^a$. During decoding, the LAS decoder rescores multiple top hypotheses from RNN-T, $\vy_r$, represented as a lattice. Specifically, we run the LAS decoder on each lattice arc in the teacher-forcing mode, with attention on $\ve^a$, to update the probability in the arc. At the end, the top output sequence with the highest probability is extracted from the rescored lattice. 

\section{Quality Improvements}
\label{sec:quality}

\subsection{Multi-domain Data \label{sec:md_data}}

Our E2E model is trained on audio-text pairs only, which is a small fraction of data compared to the trillion-word text-only data a conventional LM is trained with. Previous work \cite{Ryan19,SainathPang19} used only search utterances. To increase vocabulary and diversity of training data, we explore using more data by incorporating multi-domain utterances as described in \cite{Arun19}. These multi-domain utterances span domains of search, farfield, telephony and YouTube. All datasets are anonymized and hand-transcribed; the transcription for YouTube utterances is done in a semi-supervised fashion ~\cite{liao2013large,soltau2016neural}.


One of the issues with using multi-domain data is that each domain has different transcription conventions. For example, search data has numerics in the written-domain (e.g., \$100) while YouTube queries are often in the spoken domain (one hundred dollars). Another issue is with respect to multiple speakers. Search queries contain only one speaker per utterance, while YouTube queries contain multiple speakers. Since a main goal is to improve the quality of search queries, we explore feeding a domain-id to the E2E model as a one-hot vector, with the id being one of the 4 domains. Following work from \cite{li2018multi}, we find it adequate to only feed the domain-id to the RNN-T encoder.

\subsection{Robustness to Accents \label{sec:accent_data}}
Conventional ASR systems operate on phonemic representations of a word \cite{Jurafsky2000}. Specifically, a lexicon maps each word in the vocabulary to a few pronunciations, represented as a sequence of phonemes, and this mapping is fixed before training. This poses challenges when it comes to accents; building an English recognizer that is accurate for American, Australian, British, Canadian, Indian, and Irish English variants is challenging because of phonetic variations. 

Attempting to solve these issues by merging the phoneme sets is difficult. Using a lexicon with an on-device E2E system significantly increases the memory footprint, since the size of the lexicon can be upwards of 0.5 GB \cite{CC18}. In addition, the increased number of phonemes causes confusion and creates data sparsity problems. Finally, decisions regarding the phoneme set and the pronunciations of a word are not made directly from data.


Instead, our E2E model directly predicts word pieces. The model itself decides how to handle pronunciation and phonetic variations based on data. Its size is fixed regardless of the number of variants. As a simple strategy to improve robustness to different accents, we explore including additional training data from different English-accented locales, using the same data as described in \cite{li2018multi}. Specifically, we use data from Australia, New-Zealand, United Kingdom, Ireland, India, Kenya, Nigeria and South Africa. We down-weight the data proportion from these locales by a factor of $0.125$ during training. This number was chosen empirically to be the largest value that did not degrade performance on the American English set.


Spelling conventions vary from one variant of English to another. Since our training data was transcribed using the spelling convention of the locale, using the raw transcript can potentially cause unnecessary confusion during training. The E2E model might try to learn to detect the accent in order to decide which spelling convention to use, thus degrading robustness. Instead, we used VarCon \cite{VarCon} to convert the transcripts to the American spelling convention. For each word in the target, we use VarCon's many-to-one mapping for conversion, and then use the converted sentence as a target. In addition, during inference when evaluating accented test sets, we convert all reference transcipts to the American spelling as well.

\subsection{Learning Rates}

Our past work has explored using an exponentially-decaying learning rate when training both RNN-T and LAS~\cite{Ryan19,SainathPang19}. Given the increased amount of multi-domain training data compared to search-only data, we explore using a constant learning rate. To help the model converge, we maintain an exponential moving average (EMA) \cite{Polyak92} of the weights during training and use the EMA weights for evaluation.

\section{Latency Improvements}
\label{sec:latency}



\subsection{Endpointer}

An external voice activity detector (VAD)-based endpointer is often used to detect speech and filter out non-speech. It declares an end-of-query (EOQ) as soon as the VAD observes speech followed by a fixed interval of silence. EOQ-based endpointers which directly predict \eos and have been shown to improve latency \cite{Matt17}. The EOQ detector can also be folded into the E2E systems for joint endpointing and recognition by introducing a \eos token into the training target vocabulary of the RNN-T model \cite{Shuoyiin19}. During beam search decoding, \eos is a special symbol that signals the microphone should be closed. Premature prediction of \eos causes deletion errors, while late prediction increases latency. 

In this work we extend the joint RNN-T endpointer (EP) model and address the above issue by applying additional early and late penalties on the \eos token. Specifically, during training for every input frame in $\vx=\{x_1, \ldots, x_T\}$ and every label $\vy=\{y_1, \ldots, y_U\}$, RNN-T computes a $U \times T$ matrix $P_{RNN-T}(\vy|\vx)$, which is used in the training loss computation. Here label $y_U$ is \eos, the last label in the sequence. We denote $t_{\eos}$ as the frame index after the last non-silence phoneme, obtained from the forced alignment of the audio with a conventional model. The RNN-T log-probability $P_{RNN-T}(y_U|\vx)$ is modified to include a penalty at each time step $t$ for predicting \eos too early or too late. $t_\text{buffer}$ gives a grace period after the reference $t_{\eos}$ before this late penalty is applied, while $\alpha_\text{early}$ and  $\alpha_\text{late}$ are scales on the early and late penalties respectively. All hyperparameters are tuned experimentally. 
\vspace{-0.05in}
\begin{align*}
\log P_{RNN-T}(y_U|x_t) \mathrel{{-}{=}} \big( & \max(0, ~\alpha_\text{early} * (t_{\eos} - t)) + \\
	                                 &  \max(0, ~\alpha_\text{late} * (t - t_{\eos} - t_\text{buffer})) \big)
\end{align*}

In this work, the RNN-T model is trained on a mix of data from different domains. This poses a challenge for the endpointer models as different applications may require different endpointing behaviors.
Endpointing aggressively for short search-like queries is preferrable, but can result in deletions for long-form transcription tasks like YouTube. Since the goal of this work is to improve the latency of search queries, we utilize the fed-in domain-id to only add the \eos token for the search queries, which addresses the latency on search queries while not affecting other domains.


\subsection{LAS Rescoring}
\label{sec:las_latency}
We apply LAS rescoring to a tree-based lattice, instead of rescoring an N-best list, for efficiency, as it avoids duplicate computation on the common prefixes between candidate sequences \cite{SainathPang19}. We further reduce the LAS latency with batch inference of the arcs when expanding each lattice branch for rescoring, as it utilizes matrix-matrix multiplication more efficiently. Furthermore, we reduce the 2nd-pass latency by offloading the computation of the additional encoder as well as the attention source keys and values to the 1st-pass in a streaming fashion, whose outputs are cached to be used in the 2nd-pass.


\section{Experimental Details}
\label{sec:experiments}

All models are trained using a 128-dimensions log-mel feature frontend \cite{Arun19}.
The features are computed using 32 msec windows with a 10 msec hop.
Features from 4 contiguous frames are stacked to form a 512 dimensional input representation,
which is further sub-sampled by a factor of 3 and passed to the model.
Following \cite{Ryan19,SainathPang19}, all LSTM layers in the model are unidirectional,
with 2,048 units and a projection layer with 640 units.
The shared encoder consists of 8 LSTM layers, with a time-reduction layer after the 2nd-layer.
The RNN-T decoder consists of a prediction network with 2 LSTM layers, and a joint network with a single feed-forward layer with 640 units. The additional LAS-specific encoder consists of 2 LSTM layers. The LAS decoder consists of multi-head attention~\cite{Vaswani17} with 4 attention heads, which is fed into 2 LSTM layers. Both decoders are trained to predict 4,096 word pieces~\cite{Schuster2012}.

The RNN-T model has 120M parameters. The additional encoder and the LAS decoder have 57M parameters. All parameters are quantized to 8-bit fixed-point, as in our previous work~\cite{Ryan19}. The total model size in memory/disk is 177MB. All models are trained in Tensorflow~\cite{AbadiAgarwalBarhamEtAl15} using the Lingvo~\cite{shen2019lingvo} toolkit on $8\times8$ Tensor Processing Units (TPU) slices with a global batch size of 4,096.

In addition to the diverse training sets described in Sec.~\ref{sec:md_data} and \ref{sec:accent_data}, multi-condition training (MTR) ~\cite{peddinti2016mtr,kim2017mtr} and random data down-sampling to 8kHz \cite{Li12} are also used to further increase data diversity. Noisy data is generated at signal-noise-ratio (SNR) from 0 to 30~dB, with an average SNR of 12~dB, and with T60 times ranging from 0 to 900 msec, averaging 500 msec. Noise segments are sampled from YouTube and daily life noisy environmental recordings. Both 8~kHz and 16~kHz versions of the data are generated, each with equal probability, to make the model robust to varying sample rates.

The main test set includes $\sim$14K Voice-search utterances (\emph{VS}) extracted from Google traffic. Additionally, we use test sets with numeric (\emph {Num}) and multi-talker interfering speech data (\emph{MT}), with $\sim$4K and $\sim$6K utterances, respectively, to test robustness of the proposed models. Accented test sets come from the following locales: Australia (en-au), United Kingdom (en-gb), India (en-in), Kenya (en-ke), Nigeria (en-ng), and South Africa (en-za), with approximately 14k, 10K, 5K, 12K, 15K and 10K utterances, respectively. All test sets are anonymized and hand-transcribed.

\section{Results}
\label{sec:results}


\subsection{Quality}
In this section, all results presented are without endpointer and LAS rescoring.

\subsubsection{Domain-ID Models}

First, we analyze the behavior of RNN-T when training with multi-domain (MD) data. Table \ref{table:md} shows the behavior on 3 datasets when training with Voice Search (VS) vs. Multi-domain data. The conventional model \cite{Golan16} ($B0$) is also listed. The table shows that while behavior on $VS$ and $MT$ improves with MD data ($E1$) compared to $E0$, performance on the numeric set degrades significantly due to the spoken-domain issue of MD data discussed in Section \ref{sec:md_data}. However, once we train with a domain-id (DI) in $E2$, performance across all 3 sets improves, and outperforms $B0$ on $Num$ and $MT$. 

\begin{table}[h!]
\centerline{
  \begin{tabular}{|c|c|c|c|c|}
    \hline
    Exp ID & Train & VS & Num & MT \\ \hline  
    B0 & Conventional & 6.3  & 13.3 & 8.4\\ \hline
    E0 & VS  & 6.8  & \textbf{10.1} &  10.4  \\ \hline 
    E1 & MD & 6.7 & 11.7 & 8.0 \\ \hline
    E2 & MD + DI & \textbf{6.6}  & 10.4 & \textbf{7.7} \\ \hline
  \end{tabular}
}
\caption{Results for multi-domain RNN-T models.}
\vspace{-0.3in}
\label{table:md}
\end{table}

\subsubsection{Robustness to Accents}

Next, we explore the behavior when including accented English data in training. Table \ref{table:enx_results} shows that $E2$ (MD+DI) degrades significantly on accented test sets compared to the baseline conventional model $B0$, which is trained with a large lexicon. $E3$, which includes accented data, improves over $B0$ on all accented sets. This demonstrates that injecting data with alternative accents helps for E2E models that are trained directly to output wordpieces, bypassing a lexicon.

\begin{table}[h!]
\centerline{
  \begin{tabular}{|c|c|c|c|}
    \hline
    Exp ID & B0 & E2 & E3 \\ \hline
    Training Data & Conventional & MD + DI & + enX \\ \hline
    VS &\textbf{6.3}  & 6.6 & 6.7 \\ \hline
    en-au & 12.1 &12.6 & \textbf{10.3} \\ \hline
    en-gb & 11.2 & 10.9& \textbf{9.1} \\ \hline
    en-in & 23.9 & 24.7 & \textbf{17.8} \\ \hline
    en-ke & \textbf{27.2} & 28.3 & \textbf{27.2} \\ \hline
    en-ng & 25.6 & 23.6 & \textbf{22.8} \\ \hline
    en-za & \textbf{14.3} & 15.7 & 14.8 \\ \hline
  \end{tabular}
}
\caption{Results including accented English data in training.}
\label{table:enx_results}
\vspace{-0.2in}
\end{table}

\subsubsection{Learning Rates}

Next, we explore performance of RNN-T when decaying the learning rate (LR) ($E3$) compared to using a constant LR ($E4$), which should have more benefits given the larger number of utterances in the MD training set. Table \ref{table:lr} shows that using a constant LR improves performance on $VS$ and $MT$ by $\sim$7\% and $\sim$8\% relative respectively, without significantly harming performance on $Num$. Note that while other types of learning-rate schedule could also help; we leave optimizing learning rate schedule further for future work.

\begin{table}[h!]
\centerline{
  \begin{tabular}{|c|c|c|c|c|}
    \hline
    Exp ID & Train & VS & Num & MT\\ \hline  
    E3 & decay LR & 6.7 & \textbf{10.4} & 7.7 \\ \hline
    E4 & const LR & \textbf{6.2}  & 10.5 & \textbf{7.1} \\ \hline
  \end{tabular}
}
\caption{Results for different learning rate schedule.}
\label{table:lr}
\vspace{-0.3in}
\end{table}

\subsection{Latency} \label{sec:result_latency}

In this section, we analyze results with the various latency improvements proposed in Section \ref{sec:latency}. The endpointer latency is measured by the median (EP50) and the 90-percentile latency (EP90).

\vspace{-0.1in}
\subsubsection{E2E Endpointer}


We first apply an external EOQ-based endpointer to the E4 RNN-T model~\cite{chang19unified}. The endpointer model and the RNN-T model are optimized independently. This degrades WER since the endpointer might cut off the decoding hypotheses when the speaker has a short pause or the ASR model is not confident and delays the outputs. We report the best operating point that balances WER and latency gains obtained via sweeping endpointer parameters during decoding~\footnote{For E2E EP, we sweep an added penalty to \eos during decoding~\cite{chang19unified}.}.
With the acoustic endpointer alone, we degrade the WER from 6.2\% (no EP) to 7.4\% to achieve a 450ms EP50 latency and 860ms EP90 latency. The joint RNN-T EP model that predicts \eos as a target in the RNN-T model training (E5) obtains a WER of 6.8\% and reduces EP50 and EP90 by 20ms and 70ms, respectively. Like \cite{Shuoyiin19}, E5 also combines EOQ for better endpointing coverage. It has a better WER and latency tradeoff than E4, which uses the acoustic EP alone.


\begin{table}[h!]
\centerline{
  \begin{tabular}{|c|c|c|c|c|c|c|}
    \hline
    Exp ID & EP & VS & EP50 & EP90 \\ \hline 
    E4 & no EP & 6.2 & N/A & N/A  \\ \hline
    E4 & EOQ & 7.4 & 450 & 860  \\ \hline  
    E5 & Joint RNN-T EP + EOQ & 6.8 &  430 &  790 \\ \hline  
  \end{tabular}
}
\caption{Results on VS with endpointer on.}
\vspace{-0.3in}
\label{table:ep}
\end{table}

\subsubsection{Second-Pass LAS Rescoring}
\label{sec:las_results}


Next, we explore adding LAS rescoring (E6), where LAS is first trained with cross-entropy and then with MWER \cite{prabhavalkar2018minimum, SainathPang19}. The RNN-T model is kept unchanged during LAS training. Table~\ref{table:2pass} shows that adding LAS for rescoring reduces WER by 10\% relative, from 6.8\% to 6.1\%, while not affecting EP latency. As a comparison, we also list the server model (B0), and will discuss this in the next section.

\begin{table}[h!]
\centerline{
  \begin{tabular}{|c|c|c|c|c|}
    \hline
    Exp ID & Train & VS & EP50 & EP90 \\ \hline 
    E5 & RNN-T & 6.8  &  430 & 790 \\ \hline
    E6 & +LAS, MWER & 6.1 & 430 & 780 \\ \hline \hline
    B0 & Conventional & 6.6  & 460  & 870 \\ \hline
  \end{tabular}
}
\caption{Results for LAS Rescoring.}
\vspace{-0.1in}
\label{table:2pass}
\end{table}

In order to show the improvement in LAS computation latency by batch inference, we benchmark the wall time for the second-pass rescoring part when we run the recognition system on 100 search utterances on a Google Pixel4 phone. Inference is run on the phone's CPU. In Table~\ref{table:las_latency}, we show that batch inference reduces both median and 90-percentile computation latency by around 32\% for LAS rescoring, achieving 97ms 90\% latency.

\begin{table}[h!]
\centerline{
  \begin{tabular}{|c|c|c|}
    \hline
    Exp ID & 50\% latency & 90\% latency \\ \hline
    E6 w/o batch inference & 86 & 145 \\ \hline
    E6 w/ batch inference & 58 & 97 \\ \hline
  \end{tabular}
}
\caption{LAS rescoring computation latency (ms).}
\vspace{-0.3in}
\label{table:las_latency}
\end{table}



\subsection{Comparison to Conventional Model}

\begin{figure}[h!]
  \centering
  \includegraphics[scale=0.26]{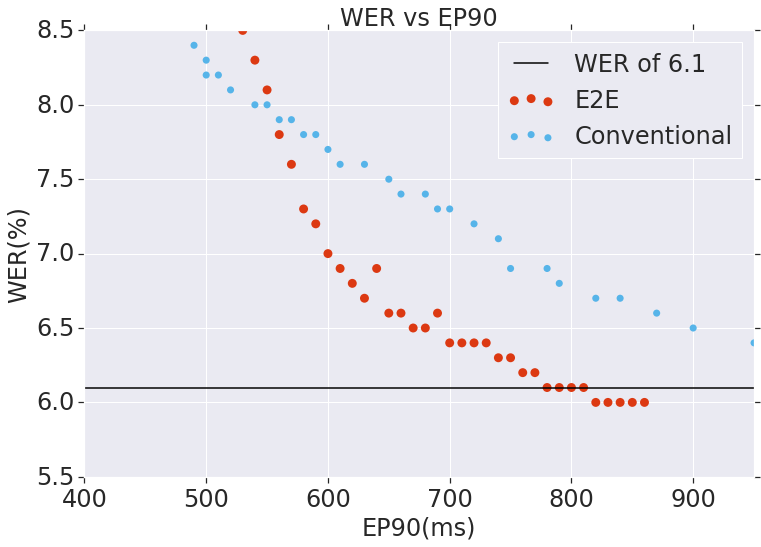}
  \caption{{WER vs EP90 for conventional model and E2E.}}
   \label{fig:server_vs_e2e_ep90}
   \vspace{-0.3in}
\end{figure}

In this section, we compare the proposed RNN-T+LAS model (0.18G in model size)
to a state-of-the-art conventional model. This model uses a low-frame-rate (LFR) acoustic model which emits context-dependent phonemes \cite{Golan16} (0.1GB), a 764k-word pronunciation model (2.2GB), a 1st-pass 5-gram language-model (4.9GB), as well as a 2nd-pass larger MaxEnt language model (80GB) \cite{Biadsy17}. Similar to how the E2E model incurs cost with a 2nd-pass LAS rescorer, the conventional model also incurs cost with the MaxEnt rescorer. We found that for voice-search traffic, the 50\% computation latency for the MaxEnt rescorer is around 2.3ms and the 90\% computation latency is around 28ms. In Figure \ref{fig:server_vs_e2e_ep90}, we compare both the WER and EP90 of the conventional and E2E models. The figure shows that for an EP90 operating point of 550ms or above, the E2E model has a better WER and EP latency tradeoff compared to the conventional model. At the operating point of matching 90\% total latency (EP90 latency + 90\% 2nd-pass rescoring computation latency) of E2E and server models, Table \ref{table:las_latency} shows E2E gives a 8\% relative improvement over conventional, while being more than 400-times smaller in size.

\bibliographystyle{IEEEbib}
\bibliography{main}
\end{document}